\documentclass[lettersize,journal]{IEEEtran}
\usepackage{amsmath,amsfonts}
\usepackage{algorithmic}
\usepackage{algorithm}
\usepackage{array}
\usepackage[caption=false,font=normalsize,labelfont=sf,textfont=sf]{subfig}
\usepackage{textcomp}
\usepackage{stfloats}
\usepackage{url}
\usepackage{verbatim}
\usepackage{graphicx}
\usepackage{cite}
\usepackage{booktabs}% http://ctan.org/pkg/booktabs - for table without border
\usepackage{siunitx}
\usepackage{gensymb}
\usepackage{mathtools}
\usepackage[dvipsnames]{xcolor}
\usepackage[
    breaklinks=true,
    colorlinks=true,
    linkcolor=blue,
    citecolor=blue,
    urlcolor=blue
]{hyperref}

\begin{document}
\bstctlcite{IEEEexample:BSTcontrol}

\title{Helping Hand: A Continuum Robotic Grasper for Traction in Single-Channel Endoscopic Submucosal Dissection}

\author{
Bendavid Maimon$^{1,*}$, 
Thompson Wong$^{1,*}$, 
Aedan Mangan$^{1}$, 
Sukjun Kim$^{1,2}$,
Michael C. Yip$^{3}$,
Tania K. Morimoto$^{1}$

\thanks{$^{*}$These authors contributed equally to this work.}
\thanks{
$^{1}$Department of Mechanical and Aerospace Engineering, University of California San Diego, La Jolla, CA, USA.
}
\thanks{
$^{2}$School of Manufacturing Systems and Networks, Arizona State University, Mesa, AZ, USA
}
\thanks{
$^{3}$Department of Electrical and Computer Engineering, University of California San Diego, La Jolla, CA, USA.
}
\thanks{This article has supplementary downloadable material (Supplementary Video S1) available, provided by the authors
}
}

\markboth{Journal of \LaTeX\ Class Files,~Vol.~14, No.~8, February~2025}%
{Shell \MakeLowercase{\textit{et al.}}: A Sample Article Using IEEEtran.cls for IEEE Journals}

\maketitle

\begin{abstract}

Endoscopic submucosal dissection (ESD) is a minimally invasive approach for removing large polyps from the colon and offers many benefits compared to other surgical approaches. However, ESD is technically very challenging to perform, leading to limited adoption by physicians. We propose the Helping Hand, a two-segment continuum robotic grasper that, together with an electrosurgical knife, enables tissue traction and cutting within a single 3.7 mm working channel.
% ---a two-segment, continuum robotic grasper---that enables cutting and tissue traction all within a single 3.7 mm working channel. 
We achieve this functionality by creating a slot at the base of the Helping Hand that lets the electrosurgical knife extend out, enabling countertraction and cutting through one instrument lumen. We evaluate the Helping Hand's workspace and its ability to provide traction forces from different starting configurations. Finally, we demonstrate the full workflow by isolating and removing tissue in an ex vivo chicken thigh, illustrating the potential for the Helping Hand to help improve ESD procedures. 

% \begin{itemize}
%     \item ESD requires controlled tissue traction and cutting within limited working-channel space.
%     \item The Helping Hand integrates a two-segment nitinol wrist, grasper, and electrosurgical knife within a single 3.7 mm working channel.
%     \item Independent robotic actuation controls wrist articulation, grasping, and knife deployment, with load cells measuring tendon tension.
%     \item Workspace and silicone phantom tests characterize positioning capability and configuration-dependent retraction forces.
%     \item Ex vivo chicken tissue resection demonstrates the intended  tissue extraction workflow.
% \end{itemize}

\end{abstract}


\section{Introduction}  
%\IEEEPARstart{P}{ointers} 
% cites:~\cite{Pimentel-Nunes2022-gw,ALES10687,tsuji2016, Xia2019-ub, Abe2020-kx, Tziatzios2020-bs}\\
Approximately 3.9\% of people in the United States will develop colorectal cancer during their lifetime~\cite{nciSEERcolorectal}. 
For patients with precancerous colorectal lesions or selected superficial cancers, endoscopic resection offers a means of removing the lesion while preserving the bowel~\cite{ferlitsch2024colorectal}.
Endoscopic mucosal resection (EMR) can capture and remove small lesions with a snare, but large lesions often must be removed in several pieces, which makes it difficult to confirm complete excision and increases the risk of recurrence~\cite{ferlitsch2024colorectal}.
% While endoscopic mucosal resection (EMR) can be performed to capture and remove small lesions with a snare, trying to remove large lesions may result in incomplete resection or unresectability. \textcolor{blue}{check which ref to use (see comment)}
%large lesions may require removal in several pieces, making it difficult to determine from the specimen whether the entire lesion was excised~\cite{ferlitsch2024colorectal}. 
In contrast, in endoscopic submucosal dissection (ESD), the physician uses an electrosurgical knife to incise the mucosa around the lesion and dissect the submucosal tissue beneath it, permitting \emph{en bloc} removal that preserves the resection margins for pathological assessment~\cite{ferlitsch2024colorectal}. %\textcolor{blue}{is it better to say something about being able to remove large polyps while  avoiding the risks and complications of major surgery?}
ESD thus allows large lesions to be removed without colectomy, which for nonmalignant polyps carries a 14\% rate of major adverse events and 0.7\% 30-day mortality~\cite{peery2018morbidity, nakamura2015potential}.

Despite the many benefits of ESD, the procedure is technically very challenging to perform. In particular, 
%A central technical challenge in colorectal ESD is 
maintaining a clear view of the tissue immediately ahead of the knife is challenging as the lesion separates from the bowel wall and the partially freed mucosal flap often obscures the plane between the lesion and the underlying muscle. Applying traction to the flap exposes this plane and places the submucosal fibers under tension, facilitating dissection~\cite{nagata2022traction,nagata2023colorectal}. The direction of traction is also critical, where a pull that initially exposes the submucosa may become less effective as the cut advances~\cite{nagata2023colorectal}. The procedural challenge is therefore to maintain traction in a direction that keeps the dissection plane both visible to the physician and accessible to the knife.

\begin{figure}[t!]
    \centering
    \includegraphics[width=0.9\linewidth]{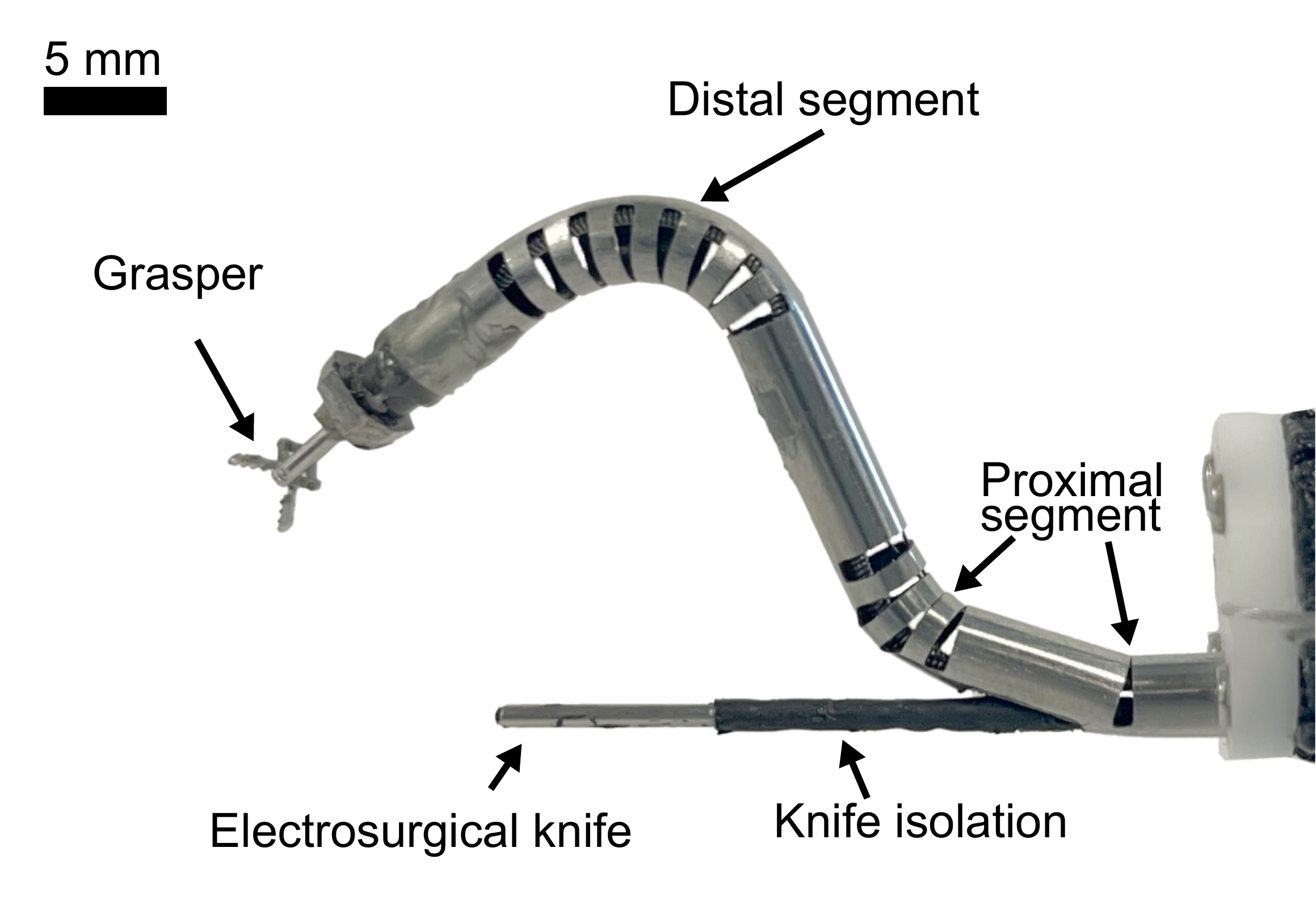}
    \vspace{-15pt}
    \caption{The Helping Hand is a two-degree-of-freedom continuum robotic grasper. It consists of two bending segments, along with a slot at the base that enables a custom electrosurgical knife to be exposed for cutting as the grasper at the tip of the Helping Hand is simultaneously used to provide tissue traction (See Supplementary Video S1).}
    % \raggedright
    \vspace{-15pt}\label{fig:helping_hand_prototype}
\end{figure}

%\subsection{Traction Devices for ESD}
% \textbf{background on other robotic tractioning systems}
% \begin{itemize}
%     \item Olympus FlexLifter~\cite{OlympusFlexLifter} provides externally mounted tissue grasping and lifting, but its interaction with the endoscope can constrain independent manipulation.
%     \item Fujifilm TRACMOTION~\cite{FujifilmTracmotion} enables articulated, rotatable grasping and real-time traction adjustment, but requires a dual-channel endoscope.
%     \item Connor et al.~\cite{Connor2026} introduce steerable sheaths for robotic manipulation through a dual-channel endoscope .
%     \item Lou et al.~\cite{Lou2026} present a soft pop-up serial robot using cable--hydraulic hybrid actuation for endoscopic manipulation.
%     \item These approaches motivate compact systems combining articulated traction and cutting without external attachments or a second working channel.
% \end{itemize}

Several devices have been proposed to maintain traction on the mucosal flap during ESD. 
%Maintaining traction on the mucosal flap during ESD requires space for both a grasper and an electrosurgical knife. 
One approach has been to mount the grasper externally, leaving the working channel available for the knife. The Olympus FlexLifter, for example, advances and retracts forceps from a fixed position on a distal cap. 
This approach results in a dependence between the angle at which the forceps approach the flap and the orientation of the endoscope~\cite{OlympusFlexLifter}. 
Lou et al. have developed an externally mounted soft pop-up robot that articulates a grasper at the tip of the endoscope~\cite{Lou2026}. %Its hybrid actuation adds control complexity because bending each joint requires coordinating cable tension for folding with hydraulic pouch inflation for unfolding. 
However, even when folded, the robot measures approximately $33 \times 32 \times 6$\,mm, adding bulk at the tip that may make it harder to navigate narrow or tortuous anatomy. 
Other robotic end caps mount to the endoscope tip and articulate instruments passed through the working channel, adding distal dexterity without a second channel~\cite{gafford2016snap}.

A second approach has been to use separate working channels for traction and cutting. The Fujifilm TRACMOTION uses an articulating, rotatable grasper to adjust the direction of traction while an electrosurgical knife passes through the second working channel~\cite{FujifilmTracmotion}. 
% Connor et al. used independently steerable sheaths for both instruments, \textcolor{blue}{can we add something else about the system here?}~\cite{Connor2026}.
Connor et al. passed two concentric push-pull steerable sheaths through the 3.7~mm and 2.8~mm channels of a standard dual-channel colonoscope, one steering forceps for countertraction and the other an electrosurgical tool, and showed that users with no prior ESD experience could complete ex vivo resections~\cite{Connor2026}.
%with a custom actuation unit that uses motor-driven lead screws and position sensors to control sheath deflection~\cite{Connor2026}. 
% Lastly, \textcolor{blue}{add in paper from BU that Aedan mentioned?}.
Gao et al. developed a bimanual telerobotic system that passes a 3.5~mm manipulator for tissue traction and a 2.5~mm manipulator for dissection through the two working channels of a commercial endoscope~\cite{gao2025transendoscopic}.
However, the externally mounted approaches add bulk at the tip of the endoscope, and the multichannel approaches require a dual-channel endoscope, whose acquisition cost limits its availability in many endoscopy units~\cite{ESDAWC}. These constraints motivate a device that provides articulated traction and knife access through a single working channel.

% \subsection{Continuum Robots}
%  % cites:~\cite{Pacheco2021-up,York2015}
%  %\cite{York2015}
% \textbf{background on continuum robots}

% \begin{itemize}

%     \item Nitinol wristed instruments offer a potential solution to ESD access and manipuation constraints through compact, and high range motion.
%     \item Their design combines force transmission and multiple degrees of freedom with low actuation complexity.
%     \item A hollow central channel enables concentric tool deployment through the wrist, integrating articulation and tool access within a single instrument.
%     \item These capabilities motivate our tool-through-wrist approach for ESD, combining repositioning and traction with tool deployment.
% \end{itemize}

\begin{figure*}[t!]
    \centering
    \includegraphics[width=\linewidth]{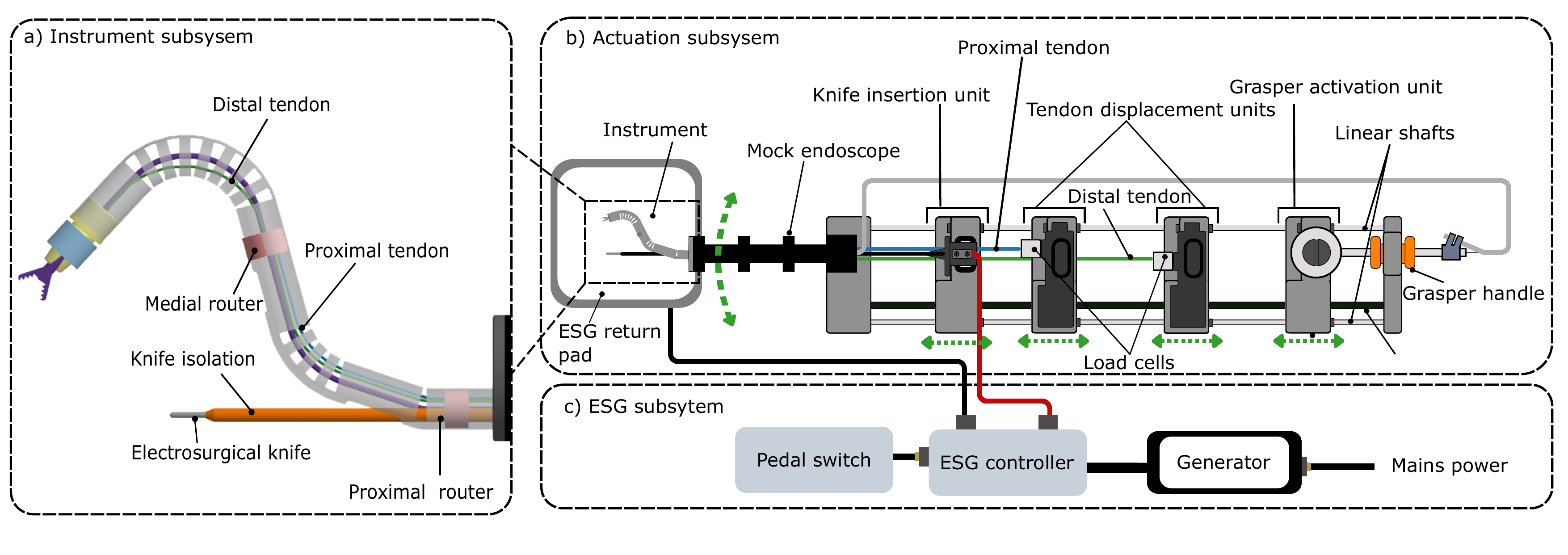}
    \vspace{-30pt}
    \caption{Overview of the Helping Hand and actuation architecture. (a) The Instrument subsystem consists of the Helping Hand and the electrosurgical knife. (b) The actuation subsystem comprises four lead-screw-driven linear motion units for proximal and distal tendon actuation, grasper operation, and knife translation. Load cells at the tendon attachments measure tension during articulation of the Helping Hand. (c) The electrosurgical generator subsystem connects to the knife and return pad through the ESG controller, with activation controlled by a pedal switch.} 
    \label{fig:HH_schematic}
\vspace{-10pt}
\end{figure*}

The contributions of this work are as follows: (1) We present the Helping Hand, a two-degree-of-freedom continuum robotic grasper that houses all the tools needed for endoscopic submucosal dissection (ESD) within a single 3.7~mm colonoscope working channel (see Fig.~\ref{fig:helping_hand_prototype}). A slot at its base, rather than the usual tip exit, lets the electrosurgical knife extend through the grasper, enabling countertraction and cutting through one instrument lumen without any externally mounted tools or a second working channel. 
    %\item We robotically actuate the wrist articulation, grasping, and knife insertion, and we extend an existing open-source continuum robot actuation platform with inline load cells that measure tension in each steering tendon during articulation.
(2) We evaluate the Helping Hand's workspace and ability to provide retraction forces. We also demonstrate the full workflow by isolating and removing tissue in an ex vivo chicken thigh benchtop model.
    To the best of our knowledge, this is the first demonstration of robotic ESD with countertraction and cutting performed through a single working channel.

% \end{figure*}
% cites:~\cite{York2015,Pacheco2021-up}
\section{System Design and Manufacturing}

The architecture of the robot, as shown in Fig.~\ref{fig:HH_schematic}, consists of two main subsystems. The \emph{instrument subsystem} consists of the Helping Hand---a two-segment, notched continuum robot with an integrated grasper---and an electrosurgical knife that passes through the Helping Hand and exits through its knife slot.  
The \emph{actuation subsystem} drives the Helping Hand and the insertion of the knife through four independent linear motion units.

% The instrument subsystem

% \begin{itemize}
%     \item The Helping Hand consists of an instrument subsystem deployed through a 3.7 mm colonoscope working channel and a back end actuation subsystem that controls wrist articulation and grasping.
%     \item The instrument assembly combines a hollow nitinol wrist, grasper, and electrosurgical knife, integrating tissue manipulation and cutting within a single instrument.
%     \item The external subsystem controls wrist articulation, grasper operation, and knife insertion, using integrated load cells to measure tendon tension on the wrist.
% \end{itemize}

\subsection{Instrument Subsystem}
\subsubsection{Helping Hand}
% \begin{itemize}
%     \item The robotic subsystem (Fig~\ref{fig:actuation_subsystem} (a)) consistsof a 3.5mm notched nitinol wrist with a 0.5mm grasper attached to the distal most point, and a central passage for electrosurgical knife deployment.
%     \item The assembly passes through the endoscope working channel, with tendons connecting the wrist and grasper to the external actuation subsystem.
%     \item Wrist articulation positions the grasper at the target tissue where the jaws close to establish traction.
%     \item The wrist then retracts the grasped tissue to expose the submucosal plane while the electrosurgical knife extends through the 1.8mm wide slit at the wrists base.
%     \item The grasper maintains countertraction during cutting, and wrist articulation adjusts tissue exposure during dissection.
% \end{itemize}

\begin{table}[b]
    \centering
    \vspace{-10pt}
    \caption{Helping Hand Parameters}
    \label{table_notch_params}
    \begin{tabular*}{\columnwidth}{@{\extracolsep{\fill}}clcc@{}}
        \toprule
        Symbol & Def. & Proximal ($p_1$, $p_2$) & Distal ($d$) \\ \midrule
           $g$   & Notch depth          & 2.8~mm  & 3.2~mm \\
           $h$   & Notch height         & 1.0~mm  & 1.0~mm \\
           $c$   & Uncut height         & 1.0~mm  & 1.0~mm \\
           $n$   & Number of notches    & 1, 4    & 8 \\
           $r_i$ & Inner radius of tube & 1.65~mm & 1.65~mm \\
           $r_o$ & Outer radius of tube & 1.75~mm & 1.75~mm \\
        \bottomrule
    \end{tabular*}
    \par\vspace{1.5mm}
    The uncut lengths at the base and tip are
    $\ell_{\mathrm{base}} = \ell_{\mathrm{tip}} = 5~\mathrm{mm}$.\par
    The gaps between notched sections are
    $\ell_{p} = 7~\mathrm{mm}$ and $\ell_{pd} = 10~\mathrm{mm}$.\par
\end{table}

\begin{figure}[t]
    \centering
    \includegraphics[width=0.9\linewidth]{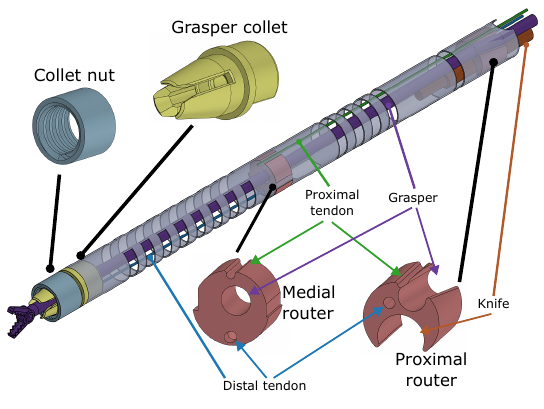}
    \vspace{-15pt}
    \caption {Exploded view of the Helping Hand, including detailed views of the medial and proximal routers, and the grasper collet.}
    \label{fig:routing}
    \vspace{-10pt}
\end{figure}

The Helping Hand is a two-segment, notched continuum robot with a slot in the proximal segment for exposing the knife (see Fig.~\ref{fig:routing}). It is fabricated from a hollow Nitinol tube~\cite{York2015, chitalia2020modeling}, and each segment is independently actuated via a tendon.  
%is fabricated from superelastic nitinol tubing for its ability to recover from the large bending deformations required to position the grasper and lift tissue. 
%Table~\ref{table_notch_params} summarizes the tube dimensions and the geometry of the notched sections.
The proximal segment consists of the knife exit notch $p_1$ and the retraction section $p_2$, with the knife slot on the opposite side of the tube.
The single notch $p_1$ sits proximal to the knife slot, so once $p_1$ bends, inserting the knife advances it out through the slot.
% and bending it allows the straight knife to become exposed when inserting.
The four notches in $p_2$ lift the grasped tissue to expose the submucosal plane.
The notches in the distal segment $d$ are oriented $180^\circ$ from those in the proximal segment, so actuating both segments produces an S bend that directs the grasper toward the polyp. Combined with insertion, this allows the tip pose to be controlled within the bending plane~\cite{mangan2025serially}.
% two notched sections, A and B, along with a slit on the opposite side of the tube. Section A consists of a single notch located proximal to the slot, enabling bending such that the straight knife can be exposed through the slot. 
% %, that when actuated, \textcolor{blue}{Aedan add in}.
% %it orients the knife slot, located between groups A and B, allowing exertion of the knife. 
% Section B consists of four notches that enable the Helping Hand to have better triangulation.
% %provide the bending motion that raises the distal half of the wrist. 
% The notches in the distal section are oriented $180^\circ$ from those in the proximal section. A separate tendon actuates the distal segment, creating an S-bend that allows the Helping Hand to bend towards and retract the polyp. 
%that directs the tip-mounted grasper toward the polyp. Once the grasper closes on the tissue, bending at section B lifts the grasped tissue to expose the submucosal plane for dissection.
Table~\ref{table_notch_params} summarizes the tube dimensions and the geometry of all notched sections.

\begin{figure}[t]
    \centering
    \includegraphics[width=\linewidth]{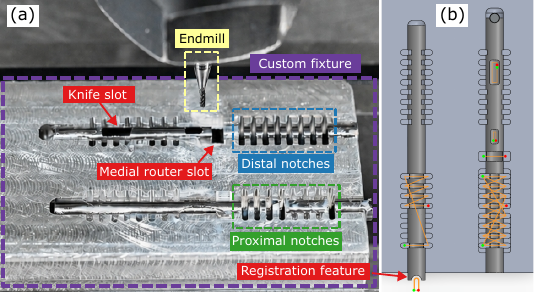}
    \caption{CNC milling setup for the Helping Hand notched tube. (a) The tube is secured in the custom fixture beneath a 1~mm end mill. (b) The CAM setup shows the programmed toolpaths in orange for Operation 1 (left) and Operation 2 (right), between which the tube is rotated $180^\circ$. Two operations on one reusable fixture produce both notch patterns with consistent alignment.}
    \vspace{-10pt}
    \label{fig:manufacturing}
\end{figure}

% \begin{itemize}
%     \item Designed a custom aluminum fixture with relief slots matching the notch geometry to securely hold the nitinol tube during CNC milling.
%     \item Incorporated a 2 mm dowel pin to ensure precise alignment when rotating the tube 180° between machining operations.
%     \item Used 1 mm carbide end mills with 3 mm and 4 mm lengths of cut to machine the proximal and distal notch patterns.
% \end{itemize}

Building on~\cite{York2015}, we developed a cost-effective and repeatable two-operation CNC milling process to fabricate the Helping Hand using a reusable aluminum fixture (Fig.~\ref{fig:manufacturing}). A groove supported the thin-walled Nitinol tube along its length, and we used Loctite Super Glue to hold it in place. Relief slots aligned with the notch pattern allowed the end mill to cut through the tube wall without contacting the fixture.
In Operation~1, we machined the proximal notches and a semicircular registration feature at the end of the tube. We then rotated the tube $180^\circ$ and seated the registration feature against a dowel pin to align the distal notch pattern for Operation~2. 
% We machined the proximal and distal notch patterns using 1~mm miniature square end mills, (Harvey Tool, 834139-C3, 836939-C3) at 9,702~rpm and a cutting feed rate of 49.3~mm/min, with a chip load of 0.00127~mm/tooth and a maximum stepdown of 0.5~mm,
% along with the knife slot and an access slot for fabrication purposes. 
Using 1~mm miniature square end mills (Harvey Tool, 834139-C3 and 836939-C3), we machined the proximal and distal notch patterns, the knife slot, and an access slot used during fabrication.
We ran the end mills at 9,702~rpm with a feed rate of 49.3~mm/min, a chip load of 0.00127~mm/tooth, and a maximum stepdown of 0.5~mm.
Finally, we released the tube in acetone and deburred the machined edges.

To ensure smooth tendon actuation and reliable operation of the grasper and knife, we designed a set of routing disks, SLA 3D printed in Grey V5 resin (Formlabs Inc.) (Fig.~\ref{fig:routing}). The proximal router establishes the initial spacing between the Nitinol tendons (0.1524~mm diameter), grasper, and knife.  
%\textcolor{blue}{add in diameter and material of tendon (0.15~mm diameter)}\
%A central design challenge was routing the 1.5 mm knife, the 1.0 mm grasper, and both steering tendons through the 3.3 mm inner diameter of the wrist while maintaining clearance during bending. 
%Two internal routing disks (Red in Fig.~\ref{fig:routing}, referred to as routers, guide these components. 
%The proximal router establishes their spacing at the wrist entrance. 
The medial router, located in the uncut length $\ell_{pd}$ between the proximal and distal segments, serves three functions. It guides the grasper shaft, anchors the proximal tendon, and redirects the distal tendon.
Redirecting the distal tendon keeps it clear of the knife path and away from the inner wall of the proximal bending segment, limiting coupling between distal tendon actuation to proximal bending.
% aligns the proximal tendon, guides the shaft of the grasper, and redirects the distal tendon. This routing keeps the distal tendon clear of the knife path and away from the inner wall of the proximal bending segment, limiting frictional contact that could couple distal tendon actuation to proximal bending. 
The grasper (Vetovation, 1~mm diameter biopsy forceps) passes through the routers and is secured to the distal end of the tube by a custom collet (Tough 1000 resin, Formlabs Inc.) and nut. 
We bonded the collet to the tube and anchored both tendons with UV curing adhesive (AA 3979, Loctite).

\subsubsection{Electrosurgical knife}

The prototype knife visible in Fig.~\ref{fig:helping_hand_prototype} fits through the 2.0~mm wide slit at the base of the Helping Hand. 
It consists of a 1.0~mm diameter stainless steel rod insulated with a silicone conformal coating (1Dfaul), which is applied evenly using a rotary fixture. 
A proximal brass shoulder is used to connect the knife to the electrosurgical generator (Force 2, Covidien Valleylab) and mounts it to its dedicated linear motion unit.
A patient return electrode (PolyHesive, Covidien Valleylab) completes the monopolar circuit through the tissue.

\subsection{Actuation Subsystem}
\label{sec_actuator}
The actuation subsystem builds on the modular system by Xu~et~al.~\cite{xu2026design}. The system consolidates all four linear motion units along a common pair of linear rails (Fig.~\ref{fig:HH_schematic}). Each linear motion unit is driven by a lead screw and independently controls one degree of freedom of the robot: proximal tendon, distal tendon, knife translation, and grasper activation. 
%The proximal and distal tendons terminate at separate tension modules. Carriage displacement of each module tensions or slacks its respective tendon to proximally or distally articulate the wrist. 
The linear motion units that actuate the tendons are fitted with inline load cells between the tendon attachment and the carriage frame, enabling real-time tension measurement during articulation of the Helping Hand. 
We extended the open source actuation platform of~\cite{xu2026design} with mechanical components for inline tension sensing and a micro-ROS~\cite{belsare2023micro} interface that connects an ESP32 microcontroller to ROS 2, and we provide both additions as open source (GitHub link to be provided upon publication).
The most distal linear motion unit controls knife advancement, and the most proximal unit actuates the grasper. We fixed the grasper trigger to the frame of the actuation unit and attached the handle to the linear motion unit, so that carriage translation opens and closes the jaws.
%The linear motion unit, clasps the grasper handle, while its trigger is fixed to the frame of the actuation unit, as shown in Fig.~\ref{fig:HH_schematic}, converting carriage translation into grasper actuation. %converted the translation of the linear motion unit into grasper actuation by
% Consistent with the routing strategy of Mangan~et~al.~\cite{mangan2025serially}, the tendons and instruments converge at the proximal end of a mock endoscope (\textcolor{blue}{add in what this was made of or any other details}) and run transverse to the knife insertion axis.
The tendons and instruments converge at the proximal to the base of the continuum segments

\section{Modeling} \label{sec_modeling}

\begin{figure}[t!]
    \centering
    \includegraphics[width=\linewidth]{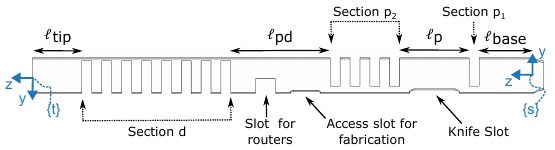}
    \caption {Side view of the Helping Hand notched tube, with the base frame $\{s\}$ at the proximal end and the tip frame $\{t\}$ at the distal end.
    One tendon actuates the proximal segment, which consists of the knife exit notch $p_1$ and retraction section $p_2$, and the second tendon actuates the distal section $d$.
    The knife slot sits between $p_1$ and $p_2$, so two tendons let a single tube expose the knife, aim the grasper, and retract tissue.
    }
    \vspace{-10pt}
    \label{fig:parametric}
\end{figure}

This section presents the forward kinematics of the Helping Hand, which map the proximal and distal tendon displacements to the pose of the tip frame $\{t\}$ relative to the base frame $\{s\}$ (Fig.~\ref{fig:parametric}).
Following~\cite{York2015}, we assumed that each notch bends with constant curvature in a steering plane and that the uncut portions of the tube remain rigid.
We then combined the individual notch transformations serially across multiple sections, similar to~\cite{mangan2025serially}.

\subsection{Notch Kinematics}
The Helping Hand consists of three notched sections (Fig.~\ref{fig:parametric}). 
The proximal segment comprises a single knife exit notch, $p_1$, and a section of four notches $p_2$, which all share the same notch geometry and steering tendon.
The distal continuum segment, $d$, has eight notches actuated by an independent tendon.
Throughout this section, the subscript $\gamma \in \{p_1, p_2, d\}$ denotes the section, and $n_\gamma$ is its number of notches ($n_{p_1}=1$, $n_{p_2}=4$, and $n_d=8$).
We assume that each tendon displacement is distributed equally among the notches it actuates.
For the proximal and distal tendon displacements $q_p$ and $q_d$, the displacement per notch is
\begin{equation}
    \Delta l_{p_1} = \Delta l_{p_2} = \frac{q_p}{n_{p_1}+n_{p_2}}
    \quad \text{and} \quad
    \Delta l_d = \frac{q_d}{n_d}.
    \label{eq:hh_notch_disp}
\end{equation}

\noindent For each section, the distance $\bar{y}_\gamma$ from the tube center to the neutral bending plane was calculated from the notch depth and tube radii in Table~\ref{table_notch_params} using the cross-sectional centroid formulation from~\cite{York2015}. 
The curvature $\kappa_\gamma$ and arc length $s_\gamma$ of a single notch are
\begin{equation}
    \kappa_\gamma \approx
    \frac{\Delta l_\gamma}
         {h_\gamma(r_i+\bar{y}_\gamma)-\bar{y}_\gamma\Delta l_\gamma}
    \quad \text{and} \quad
    s_\gamma = \frac{h_\gamma}{1+\bar{y}_\gamma\kappa_\gamma}
    \label{eq:hh_arc_parameters}
\end{equation}
\noindent where $h_\gamma$ is the notch height and $r_i$ is the inner radius of the tube~\cite{York2015}.
With bending angle $\theta_\gamma = \kappa_\gamma s_\gamma$, the transformation across notch $j$ of section $\gamma$ is
\begin{equation}
    T_{j_c}^{j+1} =
    \begin{bmatrix}
        1 & 0 & 0 & 0 \\
        0 & \cos\theta_\gamma & -\sin\theta_\gamma
          & (\cos\theta_\gamma-1)/\kappa_\gamma \\
        0 & \sin\theta_\gamma &  \cos\theta_\gamma
          & \sin\theta_\gamma/\kappa_\gamma \\
        0 & 0 & 0 & 1
    \end{bmatrix}
    \label{eq:hh_notch_transform}
\end{equation}
\noindent which reduces to a straight translation of $h_\gamma$ along the local $z$-axis as $\kappa_\gamma$ approaches zero.

\subsection{Helping Hand Forward Kinematics}
Let $T_{z,\ell}$ denote a translation by $\ell$ along the local $z$-axis and $R_{z,\phi}$ a rotation by $\phi$ about that axis. 
The transformation from the base of the first notch to the end of the last notch in section $\gamma$ is
\begin{equation}
    T_\gamma =
    \left(
        \prod_{j=0}^{n_\gamma-1} T_{j_c}^{j+1}\,T_{z,c_\gamma}
    \right) T_{z,-c_\gamma}
    \label{eq:hh_section_transform}
\end{equation}
\noindent where $c_\gamma$ is the uncut height between notches. 
The final translation $T_{z,-c_\gamma}$ ensures there are $n_\gamma - 1$ uncut heights in each section~\cite{York2015}.

The base frame $\{s\}$ is located at the proximal end of the Helping Hand, with its $z$-axis aligned with the tube and directed toward the tip (Fig.~\ref{fig:parametric}). 
The tip frame $\{t\}$ is located at the distal end of the uncut tip and is aligned with section $d$. 
Sections $p_1$ and $p_2$ share the same notch orientation, while the distal notches are oriented $180^\circ$ from the proximal notches so that section $d$ bends in the opposite direction. 
The transformation from the base frame $\{s\}$ to the tip frame $\{t\}$ is therefore
\begin{equation}
    T_{st} =
    T_{z,\ell_{\mathrm{base}}}\,
    T_{p_1}\,T_{z,\ell_{p}}\,T_{p_2}\,
    T_{z,\ell_{pd}}\,R_{z,180^\circ}\,
    T_d\,T_{z,\ell_{\mathrm{tip}}}
    \label{eq:base_to_tip}
\end{equation}

\noindent where $\ell_{\mathrm{base}}$ and $\ell_{\mathrm{tip}}$ are the uncut lengths at the base and tip, $\ell_{p}$ is the uncut length between sections $p_1$ and $p_2$ that contains the knife exit slot, and $\ell_{pd}$ is the uncut length between the proximal and distal segments that contains the medial router.

\section{Experiments and Results}

We conducted three experiments to evaluate our system. 
The first two experiments evaluated the Helping Hand alone, and the third experiment evaluated the Helping Hand together with the electrosurgical knife. 
First, we conducted a workspace characterization (Section~\ref{sec_workspace_char}) to track the Helping Hand tip during actuation of the proximal segment and distal segment and compared the results with the kinematic model. 
Second, we conducted silicone phantom traction trials (Section~\ref{sec_phantom_traction}) to measure the retraction force applied at several end effector poses. 
Third, we performed an en bloc resection (Section~\ref{sec_resection_demo}), using the Helping Hand and knife together to demonstrate the complete traction and cutting workflow in ex vivo chicken tissue. 
All experiments used the actuation platform described in Section~\ref{sec_actuator}, mounted on a lead-screw linear stage that provided the insertion axis. 
Before each trial, both steering tendons were slackened, all sensors were zeroed, and all actuation units were returned to their home positions.

\subsection{Workspace Characterization} \label{sec_workspace_char}

\begin{figure}[t!]
    \centering
    \includegraphics[width=0.8\linewidth]{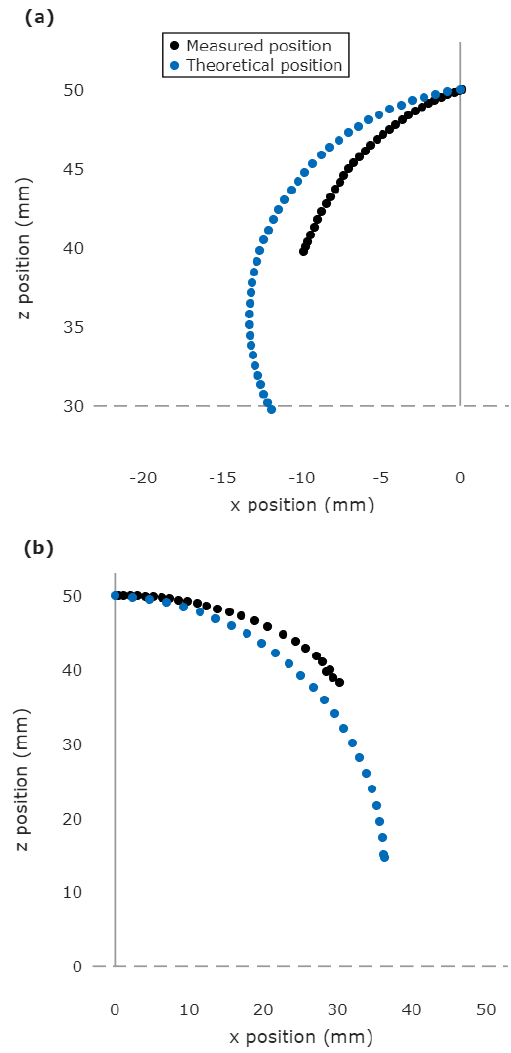}
    \caption{Measured Helping Hand tip positions (colored) and model predictions (gray) during (a) distal-only actuation and (b) proximal-only actuation, with the other tendon slack. Each tendon was displaced in 0.2~mm increments. The model captures the shape of each trajectory but overestimates tip travel, consistent with tendon elongation and friction.}
    \vspace{-10pt}
    \label{fig:workspace_results}
\end{figure}

To validate the kinematic model (Section~\ref{sec_modeling}) and characterize the reachable workspace, we tracked the tip of the Helping Hand with a field generator and an electromagnetic (EM) sensor (Northern Digital Inc.) mounted to the distal end.
% Each segment was actuated separately. 
While actuating the distal tendon, $q_d$, the proximal tendon, $q_p$, was held in slack.
The respective segment was actuated in 0.2~mm increments (Fig.~\ref{fig:workspace_results}(a)) until the notches closed fully and contacted each other.
This process was repeated for the proximal tendon, $q_p$ (Fig.~\ref{fig:workspace_results}(b)). 
Following~\cite{mangan2026contact, oliver2021concentric}, we quantified the error with the normalized root mean square error (NRMSE)
\begin{equation}
    e_{\mathrm{NRMSE}} = \frac{1}{L}\sqrt{\frac{1}{N}\sum_{i=1}^{N}\left\Vert \mathbf{p}_i-\hat{\mathbf{p}}_i \right\Vert^2}
    \label{eq:NRMSE}
\end{equation}
\noindent where $\mathbf{p}_i$ and $\hat{\mathbf{p}}_i$ are the measured and predicted tip positions at step $i$, $N$ is the number of steps, and $L$ is the length of the actuated segment. The NRMSE was 33.9\% for the proximal segment and 9.10\% for the distal segment.

As in prior notched tube designs~\cite{York2015, mangan2025serially, chitalia2020modeling, mangan2026contact}, two modeling assumptions likely accounted for much of this error. 
First, the model assumes that the full actuator displacement reaches the notched tube. 
In practice, the thin Nitinol tendons elongate under load, and the actuation system has some compliance, so the tube undergoes less displacement than the actuators command. 
Second, the model neglects friction. Because friction between the tendon and the notched tube reduces tension along the tube length, the notches bend unevenly rather than with piecewise constant curvature as the model assumes. 
%Future models could account for tendon elongation, compliance, and friction, or adopt a mechanics based formulation that relates tendon tension to notch bending. Closed loop control, using tension feedback from the existing load cells or visual feedback from the colonoscope camera, could also reduce this error.

\subsection{Silicone Phantom Traction} \label{sec_phantom_traction}

\begin{figure}[t!]
    \centering
    \includegraphics[width=0.9\linewidth]{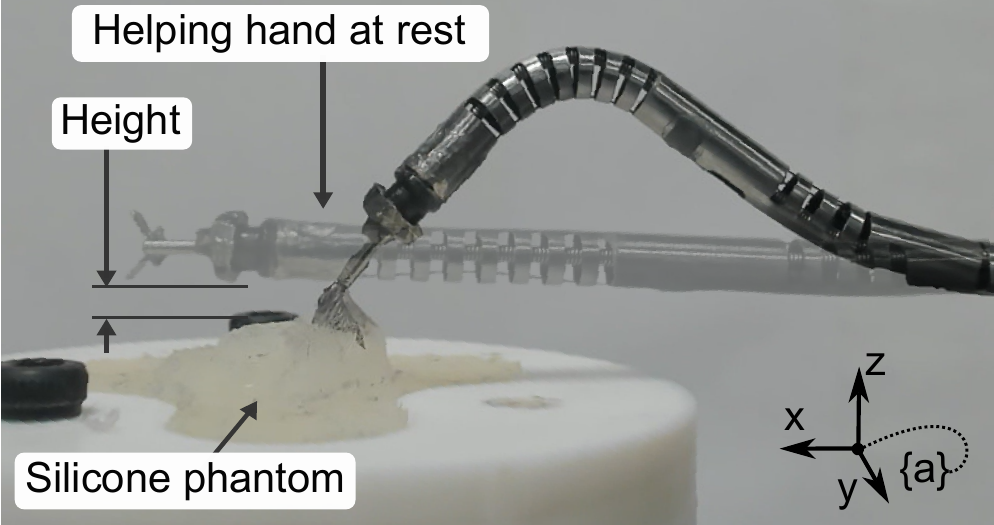}
    \caption{Traction test setup. A silicone phantom of an 18~mm sessile serrated polyp %($\sim$90th percentile in size) 
    sits on a six axis force and torque sensor below the Helping Hand. We varied the phantom height to change the Helping Hand configuration at grasping. This setup measured the retraction force on the polyp across different Helping Hand configurations.}
    \label{fig:phantom_setup}
\end{figure}

During ESD, the position of the polyp relative to the endoscope changes continuously, so the available traction may depend on the Helping Hand configuration at the time of grasping. 
To investigate this relationship, we mounted a silicone phantom of an polyp (18~mm) on a six axis force and torque sensor (Mini40, ATI Industrial Automation) (Fig.~\ref{fig:phantom_setup}). 
We recorded the tendon tensions and sensor forces and synchronized them through ROS.
From the sensor, we report two quantities. The resultant force $F_\text{res}=\sqrt{F_x^2+F_y^2+F_z^2}$ is the total magnitude of the force that the grasper applied to the phantom. 
The normal force $F_z$ is the component along the sensor $z$-axis, perpendicular to the phantom base, and represents the portion of the force that lifted the polyp away from the underlying surface.

% \begin{figure}[t!]
%     \centering
%     \includegraphics[width=\linewidth]{Figures/Draft_figs/Traction_frames_solid.png}
%     \caption{Representative phantom traction trial. 
%     The plots show the resultant and normal force on the phantom and the proximal and distal tendon tensions over time, with dashed lines marking articulation, grasping, traction, and release. 
%     The force on the phantom rose as we loaded the proximal tendon and fell when we released it.}
%     \label{fig:traction_force_response}
% \end{figure}
\begin{figure}[b!]
    \centering
    \includegraphics[width=\linewidth]{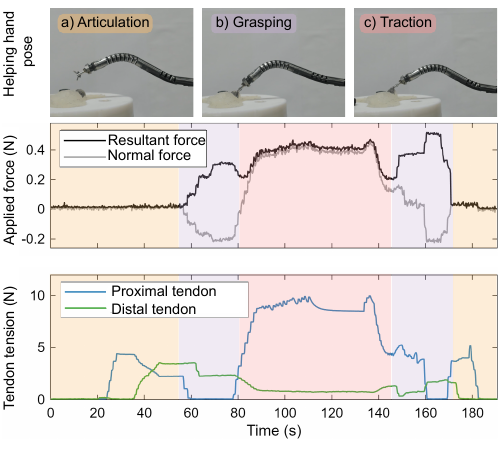}
    \vspace{-15pt}
    \caption{Representative phantom traction trial. 
    The plots show the resultant and normal force on the phantom and the proximal and distal tendon tensions over time, with dashed lines marking articulation, grasping, traction, and release. 
    The force on the phantom rose as we loaded the proximal tendon and fell when we released it.}
    \label{fig:traction_force_response}
\end{figure}
 
Figure~\ref{fig:traction_force_response} shows a representative trial. 
We first articulated the Helping Hand toward the polyp and closed the grasper on a marked point on the phantom.
We then retracted the proximal tendon until its measured tension reached 9~N. As the proximal tension increased, the force on the phantom rose. 
Finally, we returned both tendons to their initial displacements and opened the grasper, and the force on the phantom returned to zero.
To determine  single force value for each trial, we identified the samples in which the proximal tendon tension was within $9 \pm 0.25$~N and averaged $F_\text{res}$ and $F_z$ over those samples. This window captured the steady traction phase at the target tension.

\begin{table}[!t]
\caption{Force on the Phantom at 9~N Proximal Tendon Tension}
\label{table_ati_force_band_025}
  \centering
  \begin{tabular}{ccc}
    \toprule
      Height (mm) & $F_\text{res}$ (N) & $F_z$ (N) \\ \midrule
      0     & $0.331 \pm 0.015$ & $0.279 \pm 0.029$ \\
      $-4$  & $0.345 \pm 0.037$ & $0.299 \pm 0.067$ \\
      $-8$  & $0.316 \pm 0.027$ & $0.275 \pm 0.030$ \\
      $-12$ & $0.325 \pm 0.016$ & $0.243 \pm 0.037$ \\ \midrule
      All   & $0.329 \pm 0.024$ & $0.274 \pm 0.043$ \\
    \bottomrule
  \end{tabular}
  \vspace{2mm}
  \par \footnotesize Values are mean $\pm$ SD of three trials per height.
  \par Each value is averaged over proximal tension of $9 \pm 0.25$~N.
\end{table}

We repeated this procedure with the phantom at four heights, spaced 4~mm apart. 
The highest position (0~mm) placed the polyp just below the Helping Hand at rest, and the lowest position ($-12$~mm) placed it at the edge of the reachable workspace. At every height, we grasped the same marked point by translating the Helping Hand along the insertion axis. 
We performed three trials at each height.

Table~\ref{table_ati_force_band_025} summarizes the results. Across all heights, the mean resultant force was $0.33 \pm 0.02$~N and the mean normal force was $0.27 \pm 0.04$~N. 
The Helping Hand applied traction at every tested height, including a substantial component perpendicular to the phantom base, which showed that it could lift tissue throughout its workspace. 
In this experiment, we controlled only the peak proximal tendon tension in open loop. 
In future work, a more detailed force model could allow the Helping Hand to actively regulate tendon tension and maintain the desired traction while the physician cuts. 
To evaluate traction during actual cutting, we next performed an en bloc resection in ex vivo tissue.

\begin{figure*}[t!]
    \centering
    \includegraphics[width=\linewidth]{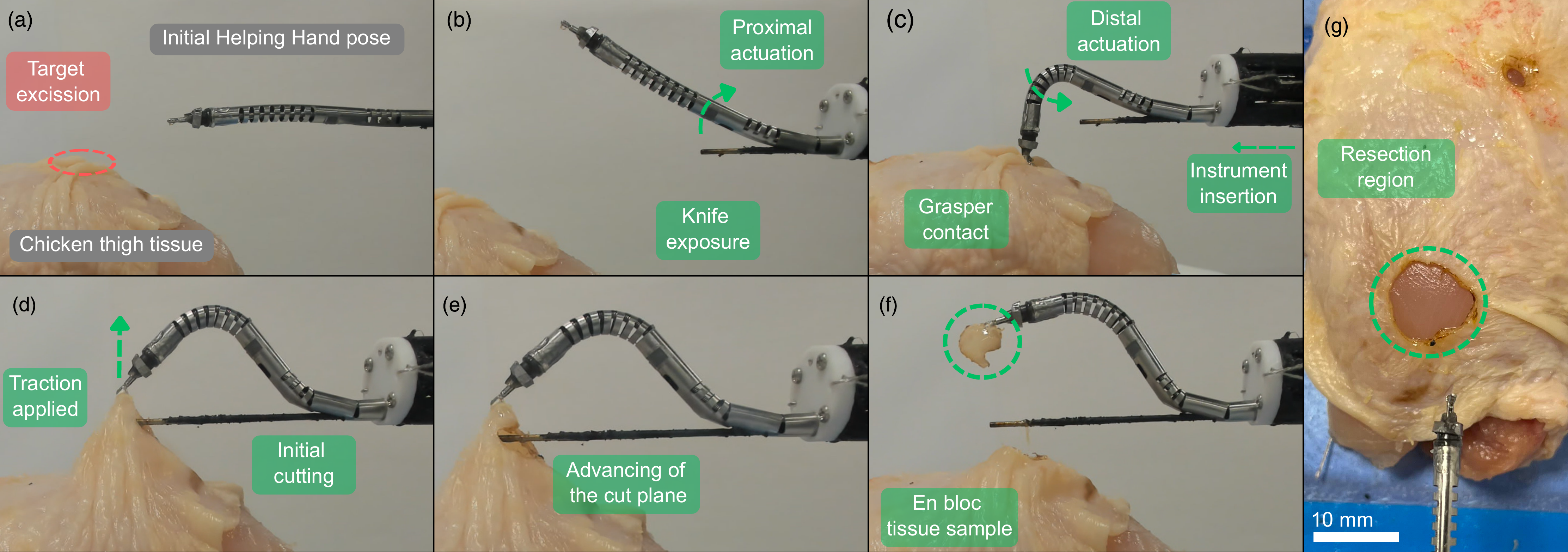}
    \caption{En bloc resection in ex vivo chicken thigh tissue with the Helping Hand and electrosurgical knife deployed from the tip of a mock endoscope. The panels show (a) insertion and alignment, (b) repositioning and knife exposure, (c) grasper positioning, (d) traction and initial incision, (e) lifting of the cut flap, (f) extraction of the grasped specimen, and (g) close up views. The Helping Hand and knife together completed grasping, traction, cutting, and retrieval through one working channel (See Supplementary Video S1).}
    \vspace{-10pt}
    \label{fig:resection_workflow}
\end{figure*}

\subsection{En Bloc Tissue Resection} \label{sec_resection_demo}

Given that en bloc removal is the main clinical advantage of ESD over other endoscopic colorectal resection techniques, the goal of this test was to evaluate whether the Helping Hand and knife could together complete the full grasp, retract, and cut workflow. 
To replicate single channel deployment, we fixed the Helping Hand to the tip of a mock endoscope, which constrained its motion to the translation it would undergo while sweeping across the dissection plane. 
We used chicken thigh tissue as the colon surrogate because its layered connective tissue approximates the thickness and compliance of the colonic wall. 
We placed the tissue on the return electrode with its top surface level with the base of the Helping Hand and connected the knife to an electrosurgical generator (ForceTriad, Valleylab).

The supplementary video and Fig.~\ref{fig:resection_workflow} illustrate the workflow. 
We routed the Helping Hand to the target, exposed the knife through the knife slot, and closed the grasper on the tissue to apply traction. 
We then made an initial incision with the knife, lifted the cut flap with the Helping Hand, and advanced the cut while manually actuating the mock endoscope until we had fully removed the target region and retrieved the specimen. 
% The excised specimen measured 27~mm in diameter (Fig.~\ref{fig:resection_workflow}(g)), and 
The full procedure took 15~min. 
These results showed that the Helping Hand and knife, deployed together through a single 3.7~mm working channel, provided both traction and cutting for en bloc resection without external tooling.

\section{Conclusion}

% This work introduced the Helping Hand, a novel continuum robotic instrument that unifies tissue traction and electrosurgical dissection into a 3.7~mm working channel of a standard single channel colonoscope. The robot achieved this by integrating a two segment notched nitinol wrist, a micrograsper, and an ESD knife into one 3.5~mm assembly. The two segment wrist providing the dextrous articulation needed for retraction of tissue, as confirmed through electromagn

In this paper, we presented the Helping Hand, a continuum robotic grasper for ESD that, together with an electrosurgical knife, enables tissue traction and cutting through the working channel of a standard single-channel colonoscope. 
We achieved this by creating a two-segment, notched continuum robot with a slot built into the base to expose the knife. 
We characterized the workspace of the Helping Hand and its ability to provide traction forces, and we demonstrated successful en bloc resection in ex vivo tissue. 
Overall, these results establish the feasibility of a continuum robotic grasper for ESD that  could operate through a single working channel.

%Existing robotic traction systems for colorectal ESD either mount external to the endoscope or require a restrictive dual channel endoscope whose cost limits availability.  

%With dexterous articulation expressed in workspace characterization, and system performance demonstrated in bench top chicken thigh models, where the system successfully dissected a 27~mm region at 29.6~mm$^2$/s while maintaining consistent retraction on the skin flap, demonstrating the complete grasp-retract-cut workflow through a single channel.

There are several directions to be explored in future work. 
First, the kinematic models could be extended to account for tendon elongation, actuation compliance, and friction, or replaced with a mechanics-based model that relates tendon tension to notch bending.
% updated to account for tendon elongation, compliance, and friction, or a new mechanics-based formulation that relates tendon tension to notch bending could be derived. 
Second, closed-loop control, using tension feedback from the existing load cells or visual feedback from the colonoscope camera, could also be incorporated. 
Finally, the Helping Hand could enable automation of portions of the ESD workflow, such as adjusting countertraction in response to forces appliaed by the knife and the tissue moving. 
% , including providing the necessary counter-traction forces based on the forces applied by the knife. 
Overall, by eliminating the need for a second working channel, an auxiliary endoscope, or intra-procedural tool exchanges, the Helping Hand could enable physicians to perform ESD using the standard colonoscopes already present in most endoscopy suites, lowering the procedural barrier that currently limits ESD adoption outside specialized centers.

%As well as automation of portions of the esd workflow such as autinomous circumfrential cutting that prevents colon perforations.

%\textcolor{blue}{moving the following from earlier section to be incorporated here} Future models could account for tendon elongation, compliance, and friction, or adopt a mechanics based formulation that relates tendon tension to notch bending. Closed loop control, using tension feedback from the existing load cells or visual feedback from the colonoscope camera, could also reduce this error.

% \clearpage
\bibliographystyle{IEEEtran}
\bibliography{biblio}

\end{document}